\pdfoutput=1

\documentclass[11pt]{article}

\usepackage{acl}

\usepackage{times}
\usepackage{latexsym}

\usepackage[T1]{fontenc}

\usepackage[utf8]{inputenc}

\usepackage{microtype}

\usepackage{cleveref}
\usepackage{pgfplots}
\usepackage{booktabs}

\definecolor{cvd8-1}{HTML}{0c2d98}
\definecolor{cvd8-2}{HTML}{78411f}
\definecolor{cvd8-3}{HTML}{2b74e2}
\definecolor{cvd8-4}{HTML}{df5f5b}
\definecolor{cvd8-5}{HTML}{3bccf6}
\definecolor{cvd8-6}{HTML}{f4cf60}

\pgfplotstableread{
Year    Hyp Ens Sen Fix Per
2015    0   0   1   -0  -0
2016	0   1   1   -1  -0
2017	0   1   3   -3  -3
2018	1   2   1   -6  -6
2019	3   3   3   -2  -6
2020	7   4   2   -11 -7
2021	1   1   2   -1  -2
}\datatable
\pgfplotsset{compat=newest}

\title{We need to talk about random seeds}

\author{Steven Bethard \\
  University of Arizona \\
  \texttt{bethard@arizona.edu} \\}

\begin{document}
\maketitle
\begin{abstract}
Modern neural network libraries all take as a hyperparameter a random seed, typically used to determine the initial state of the model parameters.
This opinion piece argues that there are some safe uses for random seeds:
as part of the hyperparameter search to select a good model,
creating an ensemble of several models,
or measuring the sensitivity of the training algorithm to the random seed hyperparameter.
It argues that some uses for random seeds are risky:
using a fixed random seed for ``replicability''
and varying only the random seed to create score distributions for performance comparison.
An analysis of 85 recent publications from the ACL Anthology finds that more than 50\% contain risky uses of random seeds.
\end{abstract}

\section{Introduction}
Modern neural network libraries all take as a hyperparameter a \textit{random seed}, a number that is used to initialize a pseudorandom number generator.
That generator is typically used to determine the initial state of model parameters, but may also affect optimization (and inference) in other ways, such as selecting which units to mask under dropout, or selecting which instances of the training data go into each minibatch during gradient descent.
Like any hyperparameter, neural network random seeds can have a large or small impact on model performance depending on the specifics of the architecture and the data.
Thus, it is important to optimize the random seed hyperparameter as we would any other hyperparameter, such as learning rate or regularization strength.

Such tuning is especially important with the pre-trained transformer architectures currently popular in NLP (BERT, \citealp{devlin-etal-2019-bert}; RoBERTa \citealp{liu2019roberta}; etc.), which are quite sensitive to their random seeds \cite{risch-krestel-2020-bagging,dodge-etal-arxiv-2020,mosbach2021on}.
Several solutions to this problem have been proposed, including
specific optimizer setups \cite{mosbach2021on},
ensemble methods \cite{risch-krestel-2020-bagging},
and explicitly tuning the random seed like other hyperparameters \cite{dodge-etal-arxiv-2020}.


The NLP community thus has some awareness of the problems that random seeds present, but it is inconsistent in its approaches to solving those problems.
The remainder of this opinion piece first presents a taxonomy of different ways that neural network random seeds are used in the NLP community, explaining which uses are safe and which are risky.
It then reviews 85 articles published in the ACL Anthology, categorizing their random seed uses based on the taxonomy.
This analysis finds that more than 50\% of the articles include risky uses of random seeds, suggesting that the NLP community still needs a broader discussion about how we approach random seeds.

\section{A taxonomy of random seed uses}
\label{section:taxonomy}
This section highlights five common uses of neural network random seeds in the NLP community, and categorizes them as either safe or risky.

\subsection{Safe use: Model selection}
\label{section:model-selection}

The random seed is a hyperparameter of a neural network architecture that determines where in the model parameter space optimization should begin.
It may also affect optimization by determining the order of minibatches in gradient descent, or through mechanisms like dropout's random sampling of unit activations.
As the random seed is a hyperparameter, it can and should be optimized just as other hyperparameters are.
Unlike some other hyperparameters, there is no intuitive explanation of why one random seed would be better or worse than another, so the typical strategy is to try a number of randomly selected seeds.
For example:
\begin{quote}
    \textit{Instead, we compensate for the inherent randomness of the network by training multiple models with randomized initializations and use as the final model the one which achieved the best performance on the validation set\ldots} 
    \cite{bjorne-salakoski-2018-biomedical}

    \textit{The test results are derived from the 1-best random seed on the validation set.} 
    \cite{kuncoro-etal-2020-syntactic}
\end{quote}

\subsection{Safe use: Ensemble creation}
\label{section:ensemble-creation}

Ensemble methods are an effective way of combining multiple machine-learning models to make better predictions \cite{rokach_ensemble-based_2010}.
A common approach to creating neural network ensembles is to train the same architecture with different random seeds, and have the resulting models vote \cite{doi:10.1142/9789812795885_0025}.
For example:
\begin{quote}
    \textit{In order to improve the stability of the RNNs, we ensemble five distinct models, each initialized with a different random seed.}
    \cite{nicolai-etal-2017-cant}

    \textit{Our model is composed of the ensemble of 8 single models. The hyperparameters and the training procedure used in each single model are the same except the random seed.}
    \cite{yang-wang-2019-blcu}
\end{quote}

\subsection{Safe use: Sensitivity analysis}
\label{section:sensitivity-analysis}

Sometimes it is useful to demonstrate how sensitive a neural network architecture is to a particular hyperparameter.
For example, \citet{Santurkar:NEURIPS:2018} shows that batch normalization makes neural network architectures less sensitive to the learning rate hyperparameter.
Similarly, it may be useful to show how sensitive neural network architectures are to their random seed hyperparameter.
For example:
\begin{quote}
    \textit{We next (§3.3) examine the expected variance in attention-produced weights by initializing multiple training sequences with different random seeds\ldots}
    \cite{wiegreffe-pinter-2019-attention}

    \textit{Our model shows a lower standard deviation on each task, which means our model is less sensitive to random seeds than other models.}
    \cite{hua-etal-2021-noise}
\end{quote}

\subsection{Risky use: Single fixed seed}
\label{section:single-fixed-seed}

NLP articles sometimes pick a single fixed random seed, claiming that this is done to improve consistency or replicability.
For example:
\begin{quote}
    \textit{An arbitrary but fixed random seed was used for each run to ensure reproducibility\ldots}
    \cite{le-fokkens-2018-neural}

    \textit{For consistency, we used the same set of hyperparameters and a fixed random seed across all experiments.}
    \cite{lin-etal-2020-cancer}
\end{quote}
Why is this risky?
First, fixing the random seed does not guarantee replicability.
For example, the tensorflow library has a history of producing different results given the same random seeds, especially on GPUs \cite{twosigma-tensorflow-2017,kanwar-etal-tensorflow-2021}.
Second, not optimizing the random seed hyperparameter has the same drawbacks as not optimizing any other hyperparameter: performance will be an underestimate of the performance the architecture is capable of with an optimized model.

What should one do instead?
The random seed should be optimized as any other hyperparameter.
\citet{dodge-etal-arxiv-2020}, for example, show that doing so leads to simpler models exceeding the published results of more complex state-of-the-art models on multiple GLUE tasks \cite{wang-etal-2018-glue}.
The space of hyperparameters explored (and thus the number of random seeds explored) can be restricted to match the availability of compute resources with techniques such as random hyperparameter search \cite{bergstra2012random} where $n$ hyperparameter settings are sampled from the space of all hyperparameter settings (with random seeds treated the same as all other hyperparameters).
In an extremely resource-limited scenario, random search might select only a single value of some hyperparameter (such as random seed), which might be acceptable given the constraints, but should probably be accompanied by an explicit acknowledgement of the risks of underestimating performance.

\subsection{Risky use: Performance comparison}
\label{section:performance-comparison}

It is a good idea to compare not just the point estimate of a single model's performance, but distributions of model performance, as comparing performance distributions results in more reliable conclusions \cite{reimers-gurevych-2017-reporting,dodge-etal-2019-show,radosavovic2020designing}.
However, it has sometimes been suggested that such distributions can be obtained by training the same architecture and varying only the random seed.
For example:
\begin{quote}
    \textit{We re-ran both implementations multiple times, each time only changing the seed value of the random number generator\ldots we observe a statistically significant difference between these two distributions\ldots}
    \cite{reimers-gurevych-2017-reporting}

    \textit{Indeed, the best approach is to stop reporting single-value results, and instead report the distribution of results from a range of seeds. Doing so allows for a fairer comparison across models\ldots}
    \cite{crane-2018-questionable}
\end{quote}
(Note the difference between \cref{section:performance-comparison,section:sensitivity-analysis}: sensitivity analysis describes only how sensitive the model is to the random seed; performance comparision makes claims about whether one model is better than another.)

Why is this risky?
If the goal is to compare the best possible model trainable from one architecture to the best possible model trainable from another architecture, as in the case of leaderboard comparisons, then varying random seeds is generating a bunch of suboptimal models for that comparison.
If the goal is to compare the family of models that result from training one architecture to the family of models that result from training another architecture, then varying only the random seed is generating a small biased slice of the family, since the family consists of the model variations across all hyperparameter settings, not just random seeds.

What should one do instead?
If the goal is to compare the best possible models trainable from different architectures, then the random seed needs to be optimized just as we would for any other hyperparameter.
It's still a good idea to compare distributions, rather than point estimates, so standard statistical techniques can be applied.
For example, bootstrap samples may be drawn from the test set, and evaluating a model on each of those samples will give a distribution over the model's expected performance \cite{dror-etal-2018-hitchhikers}.
Comparing these distributions will give a statistically sound estimate of whether the best model found for one neural network architecture outperforms the best model found for another.
If the goal is instead to compare families of models, then it makes sense to train many versions of the same architecture, but they should be sampled to vary across all hyperparameters, not just the random seed hyperparameter\footnote{Occasionally, the random seed might be the only hyperparameter, e.g., an extreme black box machine-learning scenario where the only way to get model variants is to vary the order in which training data instances are fed to the model. In such cases, it would be acceptable to vary only the random seed.}.
Comparing these distributions will give a statistically sound estimate of whether one architecture (and not just the best-optimized instance of that architecture) is better than another.

\section{Random seed uses in ACL}
\label{section:uses-in-ACL}

Having introduced both safe and risky uses of neural network random seeds, we now turn to the current state of NLP with respect to such seeds.
The following analysis is illustrative, not exhaustive, providing a conservative estimate of the prevalence of the problem.

On 29 Jun 2021, I searched the ACL Anthology for articles containing the phrases ``random seed'' and ``neural network''\footnote{\url{ https://www.aclweb.org/anthology/search/?q=\%22random+seed\%22+\%22neural+network\%22}}.
The ACL Anthology search interface returns a maximum of 10 pages of results, with 10 results per page, so I collected all 100 search results.
Non-articles (entire proceedings, author pages, supplementary material) were excluded, as were articles where the random seeds were not used to initialize a neural network (e.g., only for dataset selection).
The result was 85 articles, from publications between 2015 and 2021.

I read each of the articles and categorized its use of random seeds into one of the five purposes introduced in \cref{section:taxonomy}.
While it is conceptually possible for an article to fall into more than one category (e.g., having both ensembles and sensitivity analysis) the vast majority of articles I read fell into a single category, typically with just a single sentence where \textit{random seed} was used.
For the tiny fraction of articles where more than one category applied, since my goal only was to get a rough distribution of random seed use, I selected a ``primary'' category arbitrarily from the categories present.
A spreadsheet detailing each article reviewed, its category of random seed use, and a snippet of English text from the article justifying my assignment of that category is available\footnote{\url{https://docs.google.com/spreadsheets/d/1ab7TlW4xgOdydZJuUAQc4XQ6Us7cAQrQG8eC2OvGeaM/edit}}.

\begin{table}
    \centering
    \begin{tabular}{@{} l l @{} r @{}}
    \toprule
    Type & Purpose & Count \\
    \midrule
    Safe & Model selection & 12 \\
    Safe & Ensemble creation & 13 \\
    Safe & Sensitivity analysis & 12 \\
    \midrule
    \multicolumn{2}{@{} l}{Safe sub-total} & 37 \\ 
    \midrule
    Risky & Fixed seed & 24 \\
    Risky & Performance comparison & 24 \\
    \midrule
    \multicolumn{2}{@{} l}{Risky sub-total} & 48 \\
    \bottomrule
    \end{tabular}
    \caption{Uses of neural network random seeds for 85 ACL Anthology articles.}
    \label{table:counts}
\end{table}

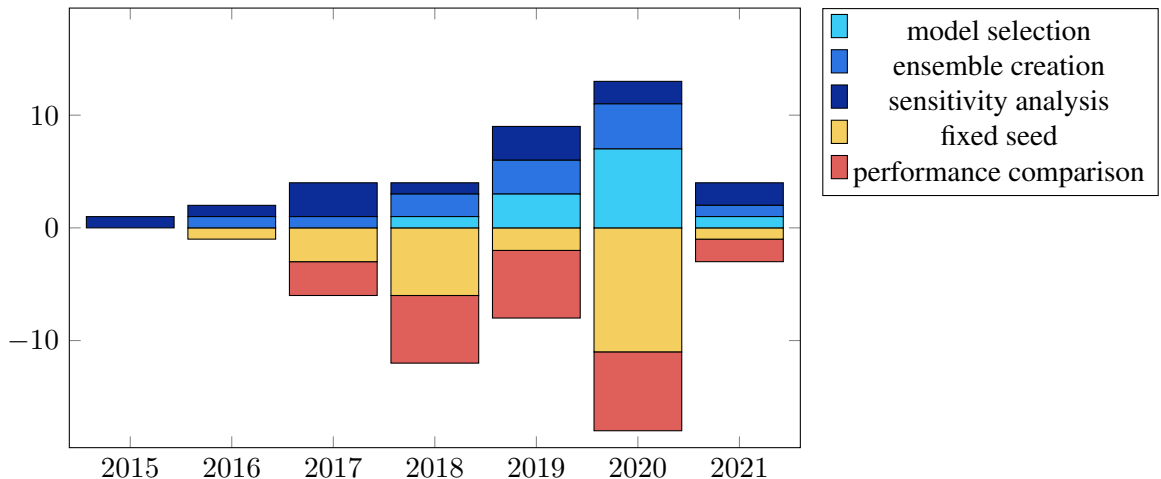
\begin{figure*}
    \centering
    \begin{tikzpicture}[]
    \begin{axis}[
        x tick label style={/pgf/number format/1000 sep=},
        ymin=-19.5, ymax=+19.5,
        ybar stacked,
        bar width=3em,
        legend pos=outer north east,
        width=0.7\textwidth,
        height=0.3\textheight,
      ]
    \addplot[fill=cvd8-5] table [y index=1] {\datatable};
    \addplot[fill=cvd8-3] table [y index=2] {\datatable};
    \addplot[fill=cvd8-1] table [y index=3] {\datatable};
    \addplot[fill=cvd8-6] table [y index=4] {\datatable};
    \addplot[fill=cvd8-4] table [y index=5] {\datatable};
    \legend{model selection,ensemble creation,sensitivity analysis,fixed seed,performance comparison}
    \end{axis}
    \end{tikzpicture}
    \caption{Uses of neural network random seeds by year for 85 ACL Anthology articles.}
    \label{figure:counts-by-year}
\end{figure*}

\Cref{table:counts} shows the distribution of articles across the different random seed purposes.
More than half of the articles (48) include a risky use of random seeds, with 24 using a single fixed seed and 24 using only random seeds to generate distributions for performance comparisons.
This suggests that NLP researchers are often using neural network random seeds without the necessary care.

One might wonder if the NLP community is getting better over time, that is, if risky uses are on the decline as NLP researchers become more familiar with neural networks research.
\Cref{figure:counts-by-year} shows that this was not the case:
though the volume of articles that matched the query varies from year to year, for most years the number of risky uses of random seeds is similar to the number of safe uses.
This suggests that NLP researchers continue to have trouble distinguishing safe from risky uses of neural network random seeds.

\section{Discussion}
We have seen that risky uses of neural network random seeds -- using only a fixed seed or generating performance distributions for model comparisons by varying only random seeds -- are still widespread within the NLP community.
The analysis in \cref{section:uses-in-ACL} is a conservative estimate of the problem.
The query used in the analysis matched articles only if they had the explicit phrases ``neural network'' and ``random seed'' both within the article.
That means the search did not return articles on neural networks where no ``random seed'' was mentioned, yet in such cases it is likely that a single fixed seed was used.
Therefore the proportion of fixed seed papers in our sample is likely an underestimate of the proportion in the true population\footnote{This query also can't address another interesting issue that is out of scope for the current article: quantifying how many articles don't optimize any hyperparameters at all.}.

How do we move the NLP community away from risky uses of neural network random seeds?
Hopefully, this article can help to start the necessary conversations, but clearly it is not an endpoint in and of itself.
Part of the responsibility must fall on mentors in the NLP community, such as university faculty and industry research leads, to ensure that they are training their mentees about these topics.
Part of the responsibility will fall on reviewers of NLP articles, who can identify misuses of neural network random seeds and flag them for revision.
And of course part of the responsibility falls on NLP authors themselves to make sure they understand the nuances of neural network hyperparameters like random seeds and the ways in which they should and should not be used.

\section{Conclusion}
This opinion piece has introduced a simple taxonomy of common uses for neural network random seeds in the NLP literature, describing three safe uses (model selection, ensemble creation, and sensitivity analysis) and two risky uses (single fixed seed and varying only the random seed to generate distributions for performance comparison). 
An analysis of 85 articles from the ACL Anthology showed that more than half of these NLP articles include risky uses of neural network random seeds.
Hopefully, highlighting this issue can help the NLP community to improve our mentorship and training and move away from risky uses of neural network random seeds in the future.

\section*{Ackmnowledgements}
Thanks to all of the anonymous reviewers; addressing their suggestions has improved this opinion piece.
However, despite interesting discussions with those reviewers, it seems that ACL Rolling Review is not conducive for papers that seek to start a discussion rather than report empirical results.
Hence, this arXiv submission.

\bibliography{anthology,custom}
\bibliographystyle{acl_natbib}

\end{document}